\documentclass[11pt,a4paper]{article}
\usepackage[hyperref]{eacl2021}
\usepackage{times}
\usepackage{latexsym}

\usepackage{todonotes}

\usepackage{amsmath,amsfonts,bm}

\def\eqref#1{equation~\ref{#1}}

\def\1{\bm{1}}

\DeclareMathAlphabet{\mathsfit}{\encodingdefault}{\sfdefault}{m}{sl}
\SetMathAlphabet{\mathsfit}{bold}{\encodingdefault}{\sfdefault}{bx}{n}

\usepackage{xspace}

\usepackage{booktabs,arydshln}

\usepackage[nameinlink]{cleveref}
\crefformat{section}{\S#2#1#3} 
\crefformat{subsection}{\S#2#1#3}
\crefformat{subsubsection}{\S#2#1#3}

\def\adl@drawiv#1#2#3{%
        \hskip.5\tabcolsep
        \xleaders#3{#2.5\@tempdimb #1{1}#2.5\@tempdimb}%
                #2\z@ plus1fil minus1fil\relax
        \hskip.5\tabcolsep}
\newcommand{\cdashlinelr}[1]{%
  \noalign{\vskip\aboverulesep
           \global\let\@dashdrawstore\adl@draw
           \global\let\adl@draw\adl@drawiv}
  \cdashline{#1}
  \noalign{\global\let\adl@draw\@dashdrawstore
           \vskip\belowrulesep}}

\newcommand{\egrid}{{\sc {EGrid}}}
\newcommand{\negrid}{{\sc {NeuralEGrid}}}
\newcommand{\lexnegrid}{{\sc LexNeuEGrid}}
\newcommand{\unimodel}{{\sc UnifiedModel}}
\newcommand{\transmodel}{{\sc TransModel}}

\newcommand{\ie}{{\em i.e.,}\xspace}
\newcommand{\eg}{{\em e.g.,}\xspace}

\newcommand{\Na}{({\em a})~}
\newcommand{\Nb}{({\em b})~}
\newcommand{\Nc}{({\em c})~}
\newcommand{\Nd}{({\em d})~}

\usepackage{url}

\usepackage{hyperref}
\usepackage{graphicx}
\usepackage{amssymb}
\usepackage{comment}
\usepackage{enumitem}
\usepackage{slashbox}
\usepackage{caption}
\usepackage{subcaption}
\usepackage{booktabs}
\usepackage{tikz}
\usetikzlibrary{trees}
\usepackage{subfiles}
\usepackage[export]{adjustbox}
\usepackage{tabu}

\usepackage{cleveref}
\crefformat{section}{\S#2#1#3} 
\crefformat{subsection}{\S#2#1#3}
\crefformat{subsubsection}{\S#2#1#3}

\usepackage{microtype}

\aclfinalcopy 

\title{Rethinking Coherence Modeling: Synthetic vs. Downstream Tasks}

\author{Tasnim Mohiuddin$^*$$^\P$, Prathyusha Jwalapuram$^*$$^\P$, Xiang Lin$^*$$^\P$, \and Shafiq Joty \thanks{*Equal contribution} $^\P$$^\dagger$ \\
$^\P$Nanyang Technological University, Singapore \\
$^\dagger$Salesforce Research Asia, Singapore \\
{\tt \{mohi0004, jwal0001, linx0057, srjoty\}@ntu.edu.sg}
\\}

\date{}

\begin{document}
\maketitle
\begin{abstract}
Although coherence modeling has come a long way in developing novel models, their evaluation on downstream applications {for which they are purportedly developed} has largely been neglected. With the advancements made by neural approaches in applications such as machine translation (MT), summarization and dialog systems, the need for coherence evaluation of these tasks is now more crucial than ever. However, coherence models are typically evaluated only on synthetic tasks, which may not be representative of their performance in downstream applications. To investigate how representative the synthetic tasks are of downstream use cases, we conduct experiments on benchmarking  well-known traditional and neural coherence models on synthetic sentence ordering tasks, and contrast this with their performance on three downstream applications: coherence evaluation for MT and summarization, and next utterance prediction in retrieval-based dialog. Our results demonstrate a weak correlation between the model performances in the synthetic tasks and the downstream applications, {motivating alternate training and evaluation methods for coherence models.\footnote{Code and data used for evaluation available at \url{https://ntunlpsg.github.io/project/coherence/coh-eval/}}}

\end{abstract}

\newcommand{\fix}{\marginpar{FIX}}
\newcommand{\new}{\marginpar{NEW}}

\section{Introduction and Related Work} 
\label{sec:intro}

Coherence is an important aspect of discourse that distinguishes a well-written text from a poorly-written one that is difficult to comprehend \cite{Halliday76}. Computational models that can assess coherence have applications in text generation and ranking, such as summarization, machine translation, {essay scoring} and dialog systems. 

Researchers have proposed a number of formal theories of discourse coherence, which have inspired the development of many coherence models -- both traditional and neural ones. Inspired by the Centering Theory \cite{Grosz:1995}, the \emph{entity} based local models \cite{Barzilay:2008,Elsner:2011} formulate coherence in terms of {syntactic roles (\eg\ subject, object)} of entities in nearby sentences. Another branch of models \cite{Pitler:2008,Lin:2011,Feng:2014} use \emph{coherence relations} between adjacent sentences to model local coherence, inspired by the discourse structure theories of \citet{Mann88} and \citet{Webber04}. Other traditional methods include \emph{word co-occurrence} based {local} models \cite{Soricut:2006}, \emph{topic} based {global} models \cite{barzilay-lee-2004,elsner-etal-2007-unified}, and \emph{syntax} based local and global models \cite{Louis:2012:CMB}.

Despite continuous research efforts in developing novel coherence models, their  usefulness in downstream applications has largely been ignored. They have been evaluated in mainly two ways. The most common approach has been to evaluate them on \emph{synthetic} discrimination tasks that involve identifying the right order of the sentences at the local and global levels \cite{Barzilay:2008,Elsner:2011,moon-et-al-emnlp-19}. The other (rather infrequent) way has been to assess the impact of coherence score as an additional feature in downstream tasks like \emph{readability assessment} and \emph{essay scoring}  \cite{Barzilay:2008,mesgar-strube-2018-neural}.  But since the concept of coherence goes beyond these constrained tasks and domains, so should the models.

{Given the recent advances in neural NLP methods, with claims of reaching human parity in machine translation \cite{Hassan2018AchievingHP}, fluency in summarization \cite{Liu2017GenerativeAN, celikyilmaz-etal-2018-deep}, or context-consistent response generation \cite{zhang2019dialogpt,hosseiniasl2020simple}, coherence {modeling} of machine-generated texts, particularly at a document-level, is now more crucial than ever \cite{Lubli2018HasMT, sharma2019entity}.} Traditional task-specific evaluation methods (\eg\ BLEU, ROUGE) may not be an accurate reflection of their real-world performance in terms of readability \cite{Paulus2017ADR, reiter2018structured}. However, it is unclear if existing coherence models are capable of this task, since their performance on downstream applications is rarely studied, {even though that is one of the main motivations for their development.}

Our main goal in this work is to assess the performance of the existing coherence models not only on standard, challenging synthetic tasks like global and local discrimination, but more importantly on real downstream text generation problems. Specifically, we investigate the performance of coherence models in three different settings:

\begin{itemize}[leftmargin=*, itemsep=0.3pt]
    \item Traditional synthetic tasks involving {discrimination of real documents from their permutations.}
    \item Coherence evaluation for machine translations and system-generated extractive and abstractive summaries, which are more representative of real-world use cases for coherence models.
    \item Next utterance ranking for dialogs, which is a downstream application similar to the synthetic task of insertion, but uses conversational data from DSTC 8 \cite{kim-dstc-2019}.
\end{itemize}

We show through experiments that there is only a slight correlation between model performances on synthetic tasks and the {real-world use cases}. Although models perform strongly in the synthetic tasks, they show poor performance and low correlations with human judgments on distinguishing coherent machine translations and {system-generated} summaries from incoherent ones. They also fail to perform well on the next utterance ranking task, {which is similar to the synthetic task of insertion  \cite{Elsner:2011}}, even if re-trained with task-specific data. 

However, we show that re-training the coherence models with task-specific data for machine translation evaluation leads to improved results and agreements with human judgments. This leads us to conclude that there is a possible mismatch in the task setting that is used to train coherence models. Models trained on traditional synthetic tasks do not seem to be learning features that are useful for downstream applications. We hope that our results will motivate the broadening of the standard of coherence model evaluations to include more downstream tasks, and also motivate the redesigning of the training paradigm for coherence models.

\section{Coherence Models} \label{sec:coh-models}

{Advancements in deep learning have inspired researchers to neuralize many of the traditional models. \citet{li-hovy:EMNLP20142} model syntax and inter-sentence relations using a recurrent sentence encoder followed by a fully-connected layer. In a follow-up work, \citet{li-jurafsky:2017} use generative models to incorporate global topic information with an encoder-decoder architecture. \citet{joty-etal-2018-coherence} propose a neural entity grid model using convolutions over distributed representations of entity transitions. \citet{mesgar-strube-2018-neural} model change patterns of salient semantic information between sentences. {\citet{xu-etal-2019-cross} propose a local discriminative  model that retains the advantages of generative models and uses a smaller negative sampling space that can learn against incorrect orderings.}  \citet{moon-et-al-emnlp-19} propose a unified  model that incorporates sentence syntax, inter-sentence coherence relations, and global topic structures in a single {Siamese} framework.}

We benchmark the performance of five {representative} coherence models on the tasks discussed above. Our selected models comprise of both traditional and neural models. Moreover, two models are currently the state-of-the-art at the time of submission (Transferable and Unified Neural Model).

\begin{description}[leftmargin=0pt] 
\item [Entity Grid (\egrid).] \citet{Barzilay:2005,Barzilay:2008} introduced the popular entity-based model for representing and assessing text coherence motivated by the Centering Theory \cite{Grosz:1995}. This model represents a text with a two-dimensional array called an \textit{entity grid}, that captures transitions of discourse entities across sentences. These {local entity transitions} are used as deciding patterns for text coherence; a {local entity transition} of length $k$ is a sequence of $\{$S,O,X,--$\}^k$ representing grammatical roles (\textit{Subject}, \textit{Object}, \textit{Other}, and \textit{Absent}, respectively) played by an entity in $k$ consecutive sentences. The \textit{salience} of the entities, quantified by the occurrence frequency, is also incorporated to identify transitions of important entities. {\citet{Elsner:2011} improve the basic entity grid by including non-head nouns as entities (with the grammatical role X). Instead of using a coreference resolver, they match the nouns to detect coreferent entities. In our work, we consider this version of the entity grid model.}

\item [Neural Entity Grid (\negrid).] A neural version of the entity grid model  was proposed by \citet{dat-joty:2017}. The grammatical roles in the grid are converted into their distributed representations, and the entity transitions are modeled in the distributed space by performing convolutions  over it. The final coherence scores are computed from convolved features that have gone through a spatial max-pooling operation. A global, document-level pairwise loss is used to train the model.

\item [Lexicalized Neural Entity Grid.] \citet{joty-etal-2018-coherence} propose an improvement of the neural entity grid (\lexnegrid) by \textit{lexicalizing} the entity transitions using off-the-shelf word embeddings to achieve better generalization.

\item [Transferable Neural Model (\transmodel).] 

In order to generalize the coherence model across domains, \citet{xu-etal-2019-cross} propose a transferable neural model that considers coherence at a local level, taking only adjoining sentences as input. Coupled with pre-training of the sentence encoders in a generative fashion, their model demonstrates significant improvements in performance, despite being a local coherence model.

\item [Unified Neural Model (\unimodel).] 
{\citet{moon-et-al-emnlp-19} propose a unified model  that captures syntax (as a proxy of intention), discourse relations, entity attention and global topic structures. The syntax is captured by incorporating an explicit language model loss. A bi-linear layer is used to capture the inter-sentential discourse relations, while light-weight convolution is used to capture the attention and topic structures.} 
\end{description}

\section{Evaluation Tasks and Experiments} 
\label{sec:tasks}
In this section, we present the performance of the coherence models on standard synthetic tasks (\ie Global/Local Discrimination), followed by the experiments where we apply the coherence models trained on the global discrimination task to three downstream tasks (\ie, abstractive summarization, extractive summarization, and machine translation). We then present the results of the coherence models re-trained on the next utterance ranking task.

For each of the coherence models, we conducted experiments with publicly available codes from the respective authors. The three recent methods use word embeddings: \lexnegrid{}, \transmodel\ and \unimodel\ use Word2vec \cite{Mikolov.Sutskever:13}, average GloVe \cite{pennington2014glove}, and ELMo \cite{PetersELMo:2018} embeddings respectively. We use the default settings and hyperparameters suggested by the authors.

\subsection{Synthetic Tasks}
Traditionally coherence models have been evaluated mostly on synthetic tasks. For comparison with previous work, we use two {representative} synthetic tasks to compare the coherence models. 

\subsubsection{Global Discrimination.} 

Introduced by \citet{Barzilay:2008}, in this task coherence models are asked to distinguish an original (coherent) document from its incoherent renderings generated by random permutations of its sentences. We follow the same experimental setting of the Wall Street Journal ({WSJ}) news dataset as used in previous studies  \cite{Elsner:2011,moon-et-al-emnlp-19,xu-etal-2019-cross}. Similar to them, we use 20 random permutations of each document for both training and testing. Additionally, we evaluate on \emph{inverse discrimination} \cite{joty-etal-2018-coherence}, where the sentence order is reversed to create the incoherent version.

\paragraph{Setup.}

We follow the same experimental settings of the \textsc{WSJ} news dataset as used in previous works  \cite{xu-etal-2019-cross,joty-etal-2018-coherence,Elsner:2011,Feng:2014}. We use 20 random permutations of each document for both training and testing, excluding the permutations that match the original one. Table \ref{table:global_dataset} summarizes the data sets used in the global discrimination task.  We randomly select 10\% of the training set for development purposes.

\begin{table}[tb!]
\centering
	\resizebox{0.75\columnwidth}{!}{%
		\begin{tabular}{l|l|c|c}
			\toprule
			& Sections & \# Doc. & \# Pairs   \\
			\midrule
			{Train} & {00-13}  & 1,378 & 26,422 \\
			{Test}  & {14-24}  & 1,053 & 20,411 \\
			\bottomrule
		\end{tabular}
	}
	\vspace{-0.5em}
	\caption{Statistics of the \textsc{WSJ} news dataset used for the \textbf{Global discrimination} task.}
	\label{table:global_dataset}
\end{table}

\paragraph{Results.}

Table \ref{table:global-discrimination} presents the results in terms of accuracy on the two \textit{global discrimination} tasks -- the \textit{standard} and the \textit{inverse} order discrimination. We see that \unimodel\ achieves the highest accuracy on the standard order discrimination task and \transmodel\ performs the best on the Inverse  order discrimination task. The other three models use entity grids, hence they may lose the {sentence-level syntactic and semantic information}.

\begin{table}[tb!]
    \centering
	\small
	\resizebox{0.97\columnwidth}{!}{%
		\begin{tabular}{l|ccc}
			\toprule 
			 \textbf{Model} & \textbf{Emb.} & \textbf{Standard}  & {\textbf{Inverse}} \\
			\midrule 
			\egrid & {--} & 81.60 & 75.78  \\
			\negrid  & {--} &  84.36 & 83.94 \\
			\lexnegrid  & {word2vec}  &  88.51 & 88.13\\
		    \transmodel & {Avg. Glove} & 91.77 & \textbf{99.62}\\
			\unimodel & {ELMo} & \textbf{93.19} & {96.78} \\
			\bottomrule 
		\end{tabular}
	}
	\vspace{-0.5em}
	\caption{Results: Accuracies of the coherence models in the \textbf{Global {Dis}crimination} task. }
	\label{table:global-discrimination}
\end{table}

\subsubsection{Local Discrimination.} Local discrimination was  proposed by \citet{moon-et-al-emnlp-19}. In this task, two documents differ only in a local context (windows of 3 sentences). In this case, the models need to be sensitive to local changes. We use the same WSJ dataset as used by \citet{moon-et-al-emnlp-19}.

\paragraph{Setup.}

We use the same \textsc{WSJ} articles used in the global discrimination task (Table \ref{table:global_dataset}) to create our local discrimination datasets. We use the code released by \citet{moon-et-al-emnlp-19} to generate these datasets.\footnote{\href{https://github.com/taasnim/unified-coherence-model}{https://github.com/taasnim/unified-coherence-model}} Sentences within a local \textbf{window of size 3} are re-ordered to form a locally incoherent text. Only articles with \textbf{more than 10 sentences} are included in the dataset. Table \ref{table:local_dataset} summarizes the datasets. We randomly select 10\% of the training set for development purposes. 

Following \citet{moon-et-al-emnlp-19}, we create four datasets for our local discrimination task: $\mathcal{D}_{w=1}$, $\mathcal{D}_{w=2}$, $\mathcal{D}_{w=3}$ and $\mathcal{D}_{w=1,2,3}$. $\mathcal{D}_{w=1}$ contains the documents where only one randomly selected window is permuted,  $\mathcal{D}_{w=2}$ contains the documents where two randomly selected windows are permuted; $\mathcal{D}_{w=3}$ is similarly created for 3 windows. $\mathcal{D}_{w=1,2,3}$ denotes the concatenated datasets.

\begin{table}[t!]
	\resizebox{1.0\columnwidth}{!}{%
		\begin{tabular}{l|l|c|cccc}
			\toprule
			& Sections & \# Doc. & \multicolumn{4}{c}{\# Pairs}   \\
			&		   &		& $\mathcal{D}_{w=1}$ & $\mathcal{D}_{w=2}$ & $\mathcal{D}_{w=3}$ & $\mathcal{D}_{w=1,2,3}$\\
			\midrule
			{Train} & {00-13}  & 748 & 7,890 & 12,280 & 12,440 & 32,610\\
			{Test}  & {14-24}  & 618 & 6,568 & 9,936 & 9,906 & 26,410 \\
			\bottomrule
		\end{tabular}
	}
		\vspace{-0.5em}
	\caption{Statistics on the \textsc{WSJ} news dataset used for the \textbf{Local discrimination} task. The $w$ denotes the number of permuted local windows in a document.}
	\label{table:local_dataset}
\end{table}

\paragraph{Results.}

\begin{table}[t!]
	\centering
	\setlength{\tabcolsep}{4pt}
	\scalebox{0.80}{
		\begin{tabular}{l|cccc}
			\toprule 
			\textbf{Model} &  $\mathcal{D}_{w=1,2,3}$ &  $\mathcal{D}_{w=1}$ &  $\mathcal{D}_{w=2}$ &  $\mathcal{D}_{w=3}$ \\
			\midrule
			\egrid & {59.78} & {53.89} & {60.43} & {63.04} \\
			\negrid  & 57.49 & 56.74 & 57.11 & 60.0 \\

			\lexnegrid  &  56.65 & 58.21 & 58.95 & 58.42\\
			\transmodel &  66.87  & 66.25 & 67.95 & 65.52 \\
			\unimodel &  \textbf{77.07} & \textbf{67.29} & \textbf{76.12} & \textbf{81.23}  \\
			\bottomrule 
		\end{tabular}
	}

	\caption{Results: Accuracies of the models in the \textbf{Local {Dis}crimination task}.} 
	\label{table:local-discrimination} 
\end{table}

From Table \ref{table:local-discrimination}, we see that the \unimodel\ achieves the highest accuracy on all four datasets. {A possible reason could be the loss function it  uses to train the model}. Unlike other models, \unimodel\ uses an adaptive pairwise ranking loss which does not penalize the locally coherent sentences. In the local discrimination task, the difference between positive and negative examples is small; {they differ only in 1-3 windows, while the other parts are locally coherent.} \unimodel{}'s loss function can model  this better.

\subsection{Coherence Evaluation Tasks}

We evaluate the coherence models trained on the global discrimination task on two downstream tasks: machine translation (MT) and summarization coherence evaluation. Note that both the MT and summarization data are from the same domain (news) as the original WSJ training data.

\subsubsection{Machine Translation Evaluation}
The outputs of neural machine translation (NMT) systems have been shown to be more fluent than their phrase-based predecessors \cite{Castilho2017IsNM}. However, recent studies have shown that there is a statistically strong preference for human translations in terms of both adequacy and fluency at a document level \cite{Lubli2018HasMT, CUBBITT}.

\citet{smith-etal-2016-trouble} evaluated traditional (non-neural) coherence models to see if they can distinguish a reference from a system translated document, and reported very  low accuracy. However, the situation has changed with the advancements of neural models; today's coherence models are claimed to be much more accurate. 

Our goal therefore is to evaluate the coherence models on how well they can judge the coherence of MT outputs at the document level. To do this, we use the system translations released by the annual Workshop (now Conference) on Machine Translation (WMT) through the years 2017 and 2018. At a document level, reference (human) translations have been shown to be more coherent than MT outputs \cite{Smith2015APF, smith-etal-2016-trouble, Lubli2018HasMT}. Therefore, we evaluate the performance of the coherence models based on their accuracy of scoring the reference (document) higher than the system translation (document).

\begin{figure}[t!]
    \centering
    \includegraphics[scale=0.12]{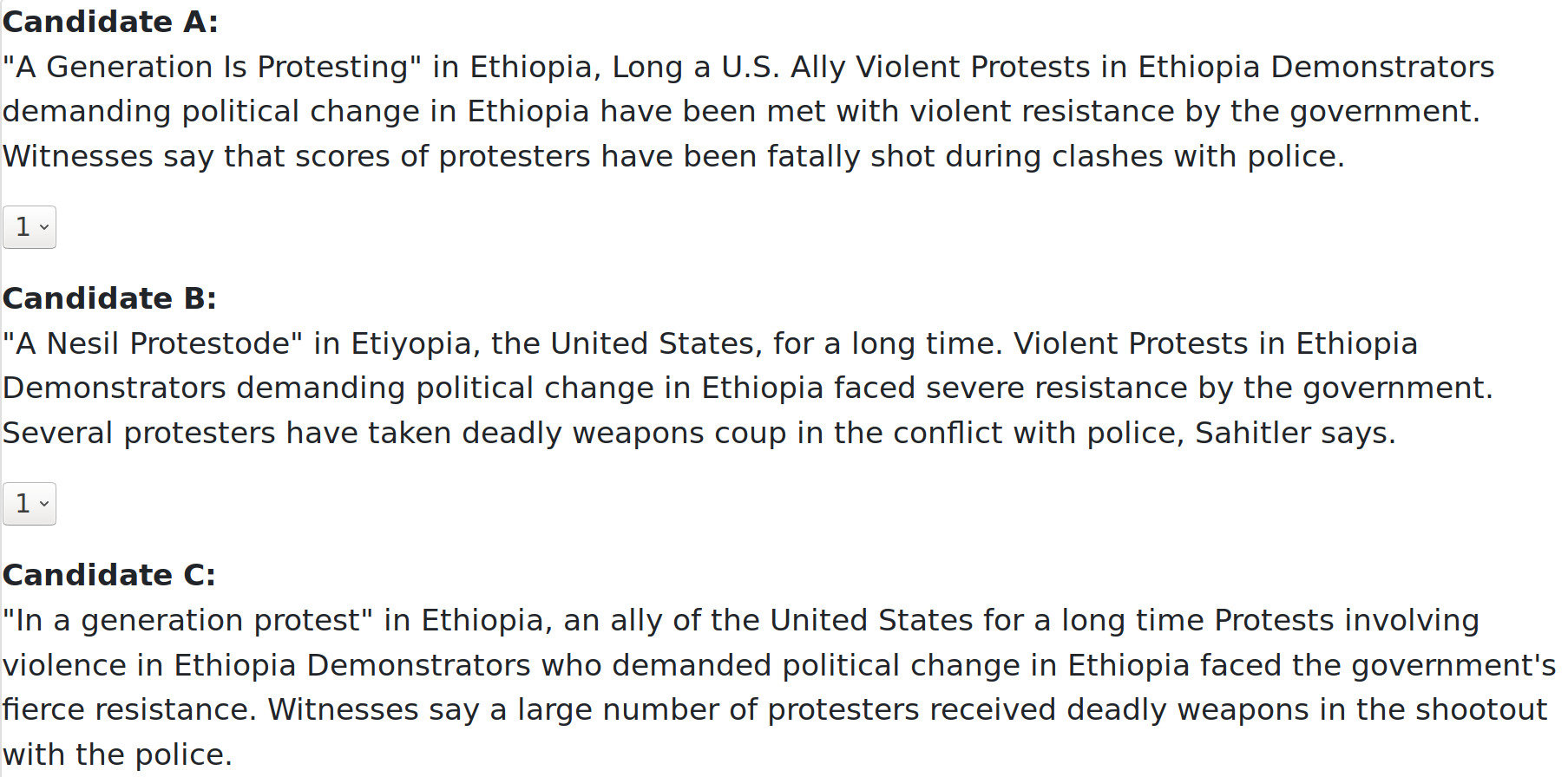}
    \caption{User study interface for coherence ranking.}
    \label{fig:coh_rank}
\end{figure}

We also obtain rankings given by humans in a user study. Fig. \ref{fig:coh_rank} shows the layout of the  study, where participants were shown four sentences from three candidate translations of the same source text and asked to rank them against each other.  One of the given translations is the reference, used as a control, and to validate our assumption that the reference is more coherent than the system translations. 3 participants annotated 100 such samples.

Participants chose the reference as more coherent with an agreement of \textbf{0.84}, confirming our assumption.\footnote{{Traditional correlation measures such as Cohen's Kappa are not robust to skewed distributions of annotations, which was an issue here since the annotators were always more likely to choose the reference as better. Thus, we report the more appropriate Gwet's AC1/gamma coefficient \cite{gwetac1}, which controls for this.}} We evaluate the system translations by producing a ranking between the different translations of the same source text. To do this, we first obtain scores from the coherence models for the reference and each of the corresponding system translations. Then, we \textit{normalize} the scores of the system translations by {subtracting them from score of the reference}. These {normalized} coherence scores are used to rank the system translations, which are then used to calculate agreements.
 
\label{subsec:mt_with_wsj}
\paragraph{Setup.} We use the reference and the system translations provided by WMT2017-2018 as our test data, under the assumption that the reference translations are more coherent than the system translations. This results in a testset of 20,680 reference-system translation document-pairs. 

\paragraph{Results.} We report the accuracy of the coherence models trained on the global discrimination task in distinguishing the more coherent reference text from the less coherent system translations in Table~\ref{tab:coh_mt_wsj_results}. We can see that most models perform worse than a random baseline of 50\%, showing that their training on the global discrimination task is not helpful in detecting coherence quality in MT text. The difference in performance is particularly glaring for the  \transmodel\ and the \unimodel{}, both of which have over 90\% accuracy on the global discrimination tasks, but only manage 48.67\% and 43.36\% on this task respectively.

We also report the agreement with human rankings on the study data in Table~\ref{tab:coh_mt_wsj_results}. Overall, only \egrid\ has good agreement with human rankings, with all other models doing similarly poorly.\footnote{Note that the study data is different from the test data, so the accuracies and agreements may not correlate.}

\begin{table}[t!]
    \centering
    \scalebox{0.85}{\begin{tabular}{l|c|c}
    \toprule
    \textbf{Model} & \textbf{Acc. (\%)} & \textbf{AC1 Agr.}  \\
    \midrule
        \egrid &  51.75 & 0.80\\
        \negrid & 54.75 & 0.77\\
        \lexnegrid & 49.34 & 0.76 \\
        \transmodel & 48.67 & 0.77 \\
        \unimodel & 43.36 & 0.78 \\
    \bottomrule  
\end{tabular}}
\vspace{-0.5em}
\caption{\textbf{Machine Translation} setting results on WMT2017-2018 data. \textbf{Acc}uracies: \% of times reference scored higher and \textbf{AC1 agr}eements for system translation rankings between annotators and models. }
    \label{tab:coh_mt_wsj_results}
\end{table}

\subsubsection{Abstractive Summarization}

Generating coherent summaries has always been a  goal in summarization  \cite{Nenkova11book}. The widely used automatic evaluation metric ROUGE \citep{rouge} measures the n-gram overlap between the generated summaries and the reference summaries at a sentence level, and thus is not sufficient for measuring coherence. \citet{summarization-critique} also recently found almost negligible correlation between ROUGE scores and human judgments on summary coherence, especially for abstractive summaries generated by recent neural summarization models.  We therefore propose to evaluate the coherence of summaries using different coherence models and measure their effectiveness on this task. 

For abstractive summarization, we use summaries from popular neural abstractive summarization systems for CNN/DM dataset \citep{hermann2015CNN, nallapati-etal-2016-abstractive}. Since abstractive  systems vary in their architectures and loss functions, they may produce very different summaries. {We run a human study to validate the rankings given by the coherence models.}

\paragraph{Setup.}

We use the CNN/DM \citep{hermann2015CNN, nallapati-etal-2016-abstractive} for this task. We collect the reference summaries from the CNN/DM testset as well as the summaries generated by the following four representative abstractive summarization systems: \Na Pointer-Generator (PG) \citep{See-17}, \Nb {\sc BertSumExtAbs} (BSEA) \citep{Liu2019TextSW}, \Nc UniLM \citep{unilm}, and \Nd SENECA \citep{sharma2019entity}.

{As discussed, we directly use the coherence models trained on the WSJ dataset for the global discrimination task.} 
The coherence models predict the scores for each system-generated summary in the testset. The scores produced by the models are then used to rank the {system-generated} summaries of the same original article. 

We conducted a user study to validate the effectiveness of the rankings produced by the coherence models. We randomly sampled 10 sets of summaries from the dataset {with each set containing four generated summaries of the same article}, 
thus resulting in $\binom{4}{2} \times 10 = 60$ pairs of {system} summaries. Two annotators were asked to rank each pair of the summaries in terms of coherence; see Appendix for the human study interface.

\paragraph{Results.}

{For the user study, the agreement between the two annotators was \textbf{0.78}, which indicates fairly reliable data.} After we obtain the rankings based on the coherence scores produced by the models, we compute the agreements between the systems and the two annotators. From the results in Table \ref{table:sumranking}, we see that \egrid\ and \lexnegrid\ show the highest agreement with human judgements. {However, despite {strong} performance in synthetic tasks, models like \unimodel\ and \transmodel\  {are unable to convert the high accuracy} into high human agreement, which demonstrates the inefficiency of current synthetic tasks.}

\begin{table}[tb!]
\centering
\scalebox{0.88}{%
\begin{tabular}{l|c|c}
\toprule 

 \bf{Models}  & \bf{Abs. Agr.}  & \bf{Ext. Agr.} \\

\midrule 
\egrid   & \bf{0.71}  & 0.52\\
\negrid  & 0.68   & \bf{0.70}\\
\lexnegrid   & \bf{0.71}  & 0.57\\
\transmodel & 0.55 & 0.38\\
\unimodel  & 0.68 & 0.35\\
\bottomrule 
\end{tabular}
}

\caption{ \textbf{Abs}tractive \textbf{Agr}eement and \textbf{Ext}ractive \textbf{Agr}eement shows the AC1 agreements for the pairwise ranking of the generated abstractive summaries and extractive summaries between two annotators and the models, respectively.}
\label{table:sumranking}
\end{table}

\subsubsection{Extractive Summarization}

For evaluating the coherence of extractive summaries, we use the dataset prepared by \citet{Barzilay:2008} for their coherence model evaluation. The dataset comes with human ratings of the summaries from the Document Understanding Conference (DUC), 2003. 

\paragraph{Setup.}

The dataset from \citet{Barzilay:2008} provides 16 sets of summaries where each set corresponds to a multi-document cluster and contains summaries generated by 5 systems and 1 human. The human ratings for these summaries based on coherence  are also available.\footnote{See Appendix for details. Rankings are available at \url{http://homepages.inf.ed.ac.uk/mlap/coherence/}}

We follow the same  experimental setup as in abstractive summarization. We use the coherence models trained on the WSJ dataset to produce scores that can be used to obtain the pairwise ranking of generated summaries.
Based on the ratings provided by \citet{Barzilay:2008}, we can generate the human pairwise rankings.

\paragraph{Results.} We present the agreements between the generated human ranking and the systems in Table \ref{table:sumranking}. We observe the same problem as in abstractive summarization that high accuracy in synthetic tasks does not lead to high human agreement in evaluating downstream {summarization} systems.

\subsection{{Task-specific Training for Dialog}}

The global and local discrimination tasks are synthetic, while the MT and summarization coherence evaluation  performance may be affected by the difference between the testing and training setup.
To control for this, we re-train and test the coherence models on a task-specific setup for next utterance ranking. This task has the advantage of being non-synthetic while providing task specific training data, but also being {similar to the synthetic task of insertion}, helping us evaluate the generalizability of the coherence model performance.

\subsubsection{Next Utterance Ranking }
The quality of a dialog depends on various conversational aspects such as engagement, coherence, coverage, conversational depth, and topical diversity \cite{see-etal-2019}. \citet{liu-etal-2016-evaluate} show that commonly used metrics such as BLEU and ROUGE show very weak or no correlation with human judgements. They also suggest using metrics that take dialog context into account. This is  particularly important as \citet{sankar-etal-2019-neural} empirically show that current neural dialog systems rarely use conversational history. We therefore propose to evaluate the usefulness of coherence models in dialog systems.

We evaluate the models on the Noetic End-to-End Response Selection Challenge II (NOESIS II), a track in the Dialog System Technology Challenges 8 (DSTC 8) \cite{kim-dstc-2019}. In this problem, each example consists of a conversational context $U = (u_1, \ldots, u_{|U|})$ and a set of potential utterances (candidates) $C=\{c_1, \ldots, c_{|C|}\}$ that may occur next in the dialog; the task is to select the correct next-utterance $r \in C$. 

This task is a nice fit for evaluating coherence models, as a good model should rank a coherent dialog higher than an incoherent one. The correct utterance along with the conversational context forms the coherent example $P = (u_1, \ldots, u_{|U|}, r)$, while other candidate utterances $c_j \in C$ with the conversational context form the incoherent examples $N = (u_1, \ldots, u_{|U|}, c_j)$. This is a considerably harder task as the difference between coherent and incoherent dialog is only the last utterance. 
We train the coherence models with these coherent ($P$) and incoherent ($N$) examples. The trained models give a score for each example based on its coherence. We then use our aforementioned assumption (coherence models should score $P$ higher than $N$) for the evaluation. This task resembles the (synthetic) \emph{insertion} task \cite{Elsner:2011} in that the goal here is to find the next correct utterance for the last position.

\begin{table}[tb!]
\centering
\resizebox{0.95\columnwidth}{!}{%
\begin{tabular}{l|c|c|c}
\toprule 
 & Train &  Dev & Test  \\
\midrule 
\multicolumn{4}{c}{\textbf{Advising dataset}} \\
\midrule 
{\# of conv.} &  50,535 & 500 & 269  \\
{\# of coh.-incoh. pairs/conv.} & 20   & 99 & 99 \\
{\# of total example pairs} & 10,10,700 & 49,500 & 26,631 \\
\midrule
\multicolumn{4}{c}{\textbf{Ubuntu dataset}} \\
\midrule
{\# of conv.} &  49,387 & 500 & 1078  \\
{\# of coh.-incoh. pairs/conv.} & 20   & 99 & 99 \\
{\# of total example pairs} & 9,87,740 & 49,500 & 1,06,722\\
\bottomrule 
\end{tabular}
}

\caption{Statistics of the refined Advising and Ubuntu datasets for the \textbf{utterance ranking} task.}
\label{table:advising-dataset}
\end{table}

\paragraph{Setup.}
We evaluated the coherence models on both datasets of the DSTC8 response selection track, \ie\ the Advising and Ubuntu datasets.\footnote{\href{https://github.com/dstc8-track2/NOESIS-II/}{https://github.com/dstc8-track2/NOESIS-II/}} The former contains two-party dialogs that simulate a discussion between a student and an academic advisor, while the latter consists of multi-party conversations extracted from the Ubuntu IRC channel \cite{acl19disentangle}. 

For a given conversational context, the goal is to select the next utterance from a candidate pool of 100 utterances, which may or may not contain the correct next utterance. We filter the datasets to suit the settings for coherence models. In our refined datasets, we exclude the conversations that have less than 7 or more than 50 utterances in the context. To ensure that we have pairwise coherent and incoherent examples, we only include the conversations that contain the correct next utterance in the candidate pool. Table \ref{table:advising-dataset} shows the statistics of our refined datasets for the utterance ranking task.

\begin{table}[tb!]
\resizebox{0.95\columnwidth}{!}{%
\begin{tabular}{l|ccccc}
\toprule 
 & \textbf{R@1} & \textbf{R@5} & \textbf{R@10} & \textbf{MRR} & \textbf{Acc.} \\
\midrule 
\multicolumn{6}{c}{\textbf{Advising dataset}} \\
\midrule
\textbf{Official Evaluation} \\
Best  & 0.564 & 0.81 & 0.88 &	0.68 & \texttt{X}\\
Median  & 0.14 &	0.37 &	0.51 & 	0.26 & \texttt{X}\\
Worst  & 0.01 & 0.05 & 0.09 &	0.05 & \texttt{X}\\
\midrule 
\textbf{Coherence Model} & & & \\
\egrid  & 0.004 &	0.03 &	0.07 &	0.04 &	47.16 \\

\negrid   &  0.057 &	0.17 &	0.23 &	0.13 &	56.15\\
\lexnegrid   & 0.046 & 0.17 &	0.26 &	0.13 &	57.66\\
\transmodel  & 0.067 & 0.20 &	0.30 & 0.14 & \textbf{66.62}  \\
\unimodel  & 0.022 &	0.06 &	0.19 & 0.11	& 54.33\\
\midrule
\multicolumn{6}{c}{\textbf{Ubuntu dataset}} \\
\midrule
\textbf{Official Evaluation} & & & \\
Best  & 0.761 &	0.96 &	0.98 & 0.85 & \texttt{X}\\
Median  & 0.55 & 0.86 & 0.93 & 0.68 & \texttt{X}\\
Worst  & 0.24 & 0.38 & 0.46 &	0.32 & \texttt{X}\\
\midrule 
\textbf{Coherence Model} & & & \\
\egrid  & 0.007	& 0.05 & 0.09 & 0.05 & 47.48 \\

\negrid   & 0.18 &	0.39 &	0.49 &	0.29	& 73.18 \\
\lexnegrid   &0.15 &	0.31 &	0.39	& 0.24 &	74.39 \\
\transmodel  & 0.045 & 0.14 & 0.26 & 0.12 & 70.94 \\
\unimodel & 0.035 &	0.17 &	0.33 &	0.13 &	\textbf{74.49}\\
\bottomrule 
\end{tabular}
}

\caption{\textbf{Utterance ranking} results for different coherence models on Advising and Ubuntu datasets. \textbf{R@k} indicates Recall@k, \texttt{X} indicates result not shared. }
\label{table:advising-results}
\end{table}

\paragraph{Results.}

Table \ref{table:advising-results} summarizes the results on the refined datasets for the utterance ranking task. {In the last column, we report the accuracy for the number of samples in which the coherence models score the positive sample higher than the negative one. All model performances are better than a random baseline, with \unimodel\ reaching 74.49\% on the Ubuntu dataset. Note, however, that because there are 100 negative samples for every positive sample, the accuracies are skewed and not representative of actual task difficulty.}

The DSTC8 challenge ranking considers the average of Recall@1, Recall@5, Recall@10 and Mean Reciprocal Rank (MRR). We report both the official evaluation results and the coherence models' performance even though the latter is tested on the refined datasets. From the results, we see that the overall performance of all the coherence models is quite poor. Despite being re-trained on task specific data, we find that coherence model performance in this task is sub-par.

\section{Task-specific Training for MT}
\label{sec:mt_retraining}

As a special use case, we report the results of re-training the coherence models using machine translation data for coherence evaluation. The aim is to investigate whether changing the usual training setup, that uses negative documents which are only small variations of the positive documents, might help coherence models learn more useful task-specific features.

\paragraph{Setup.}

Under the assumption that the reference translations are more coherent at the document level than the system translations, we train the coherence models with the reference text as the positive and the system translation as the negative document, forming a positive-negative document pair. We use the data from WMT-2011 to WMT-2015 for training (28,985 document-pairs), WMT-2016 for development (7,647 document-pairs) and the same test data (WMT-2017 to WMT-2018; 20,680 document-pairs) and study data as used for the previous experiment (\cref{subsec:mt_with_wsj}).

\paragraph{Results.}

 Table \ref{tab:coh_mt_results} reports the accuracy of the re-trained models and the results of the model ranking comparison against human rankings. Many of the models show improved performance, with the agreements increasing correspondingly. The \unimodel\ has the highest accuracy improvement by far of 34\%, improving from 43.36\% to 77.35\%. It also has the highest agreement with human rankings at 0.82. We surmise that the model's {adaptive pairwise ranking loss along with its additional language model loss boosts its performance on in-domain test data.}

\begin{table}[t!]
    \centering
    \scalebox{0.85}{\begin{tabular}{l|c|c}
    \toprule
    \textbf{Model} & \textbf{Acc. (\%)} & \textbf{AC1 Agr.}  \\
    \midrule
        \egrid & 48.74  & 0.797\\
        \negrid & 52.58 & 0.760\\
        \lexnegrid & 56.84 & 0.795\\
        \transmodel & 57.65 & 0.751 \\
        \unimodel & \textbf{77.35} & \textbf{0.828} \\
    \bottomrule  
\end{tabular}}

\caption{\textbf{Re-trained MT} setting results on WMT2017-2018 data. \textbf{Acc}uracies: \% of times reference scored higher and \textbf{AC1 agr}eements for system translation rankings between annotators and models.} 
    \label{tab:coh_mt_results}
\end{table}

\section{Discussion}
\label{sec:discussion}

{Compared to the downstream tasks of coherence evaluation in MT and extractive and abstractive summarization, the traditional global discrimination task can be considered to be a simpler task \cite{Elsner:2011}, since the difference between the positive and the negative document is a permutation/re-ordering of the sentences.} This may be rendering the models unable to learn features that are useful for downstream applications, which are likely to have other, different kinds of errors. 

On the next utterance ranking task, the models fail to generalize and perform quite poorly despite task-specific re-training. {The best model performance for the synthetic task of insertion, which is similar, also barely reaches 26\% \cite{Elsner:2011,dat-joty:2017}. This indicates that the training procedures may not be providing the right setting to learn features that are generic enough to apply to tasks in a harder setup.}

In the synthetic tasks, the models’ self-supervision comes from distinguishing an original coherent document from its incoherent renderings generated by random permutations of its sentences. This permutation-based self-supervision tries to capture document-level language properties. However, it is quite likely that this is simply a poor approximation of real-world coherence problems. Consider for example that MT systems mostly translate at the sentence-level. Consecutive sentences may lack coherence, but if two system translations of a text are compared, the translations themselves will be in the same order for both. The coherence models are not trained for such (real-world) settings.

{Another possibility is that outputs from downstream tasks have different error distributions that are captured to varying degrees by different models, since they are originally designed based on synthetic tasks. That is, models that perform very well on the permutation task might be overfitting on this task, and therefore failing to find coherence issues that are more subtle than shuffled text. Thus, we conclude that the current self-supervision for coherence modeling is not suitable for downstream coherence problems.}

When re-trained on machine translation data, most of the model performances improve, implying that a different training setting may be required to make the models applicable to actual downstream tasks. This is not apparent from the evaluation results that are usually reported, which show performances crossing the 90\% mark.

\citet{Elsner:2011-chat} show a similar lack of generalizability and applicability of coherence models to the downstream task of chat disentanglement. Our results suggest that despite nearly a decade of research since, the standard training and testing paradigm for coherence modeling continues to be inadequate in its capability to generalize to real-world use-cases and even to similar task settings, and also fails in being indicative of real-world task performance.

\section{Conclusions}
\label{sec:conclusion}
We benchmark the performance of representative traditional and neural coherence models on standard synthetic discrimination tasks, and contrast this with their performance on various downstream application tasks in NLP. We show that higher accuracies on synthetic tasks do not translate into better performance on downstream tasks. We demonstrate this for real-world tasks like MT and summarization coherence evaluation, and next utterance ranking. Our results signal a need for change in the way coherence models are typically trained and evaluated.

{Other downstream applications like coherence evaluation of language model generated text and tasks such as chat disentanglement are also good candidates for testing coherence models. It would be worthwhile to build a coherence testset that is independent of the training tasks and similar to downstream applications, which could be used by the community to test the generalization ability of their models. In future work, we also hope to investigate the possible training scenarios that will result in more generalizable coherence models which can be used for evaluating downstream tasks.  }

\bibliography{refs}
\bibliographystyle{acl_natbib}

\appendix
\section{Appendix}\label{appendix}

\subsection{Human Study Interface for Abstractive Summarization}

We show the interface of human study for abstractive summarization in Figure \ref{fig:humanintersum}. 

\subsection{Human Study for Extractive Summarization}
We briefly describe the human study for extractive summarization. The human study was conducted by \citet{Barzilay:2008}.
Coherence ratings for summaries were collected during an elicitation study by 177 unpaid native speakers of English. The annotators were asked to use a seven point-scale to rate each summary based on how coherent the summaries were without having seen the source
texts. The ratings (approximately 23 per summary) given by the subjects were averaged to provide a final rating score between 1 and 7 for each summary.

\vspace{10em}
\begin{minipage}{\textwidth}
\begin{center}
\includegraphics[width=0.99\textwidth]{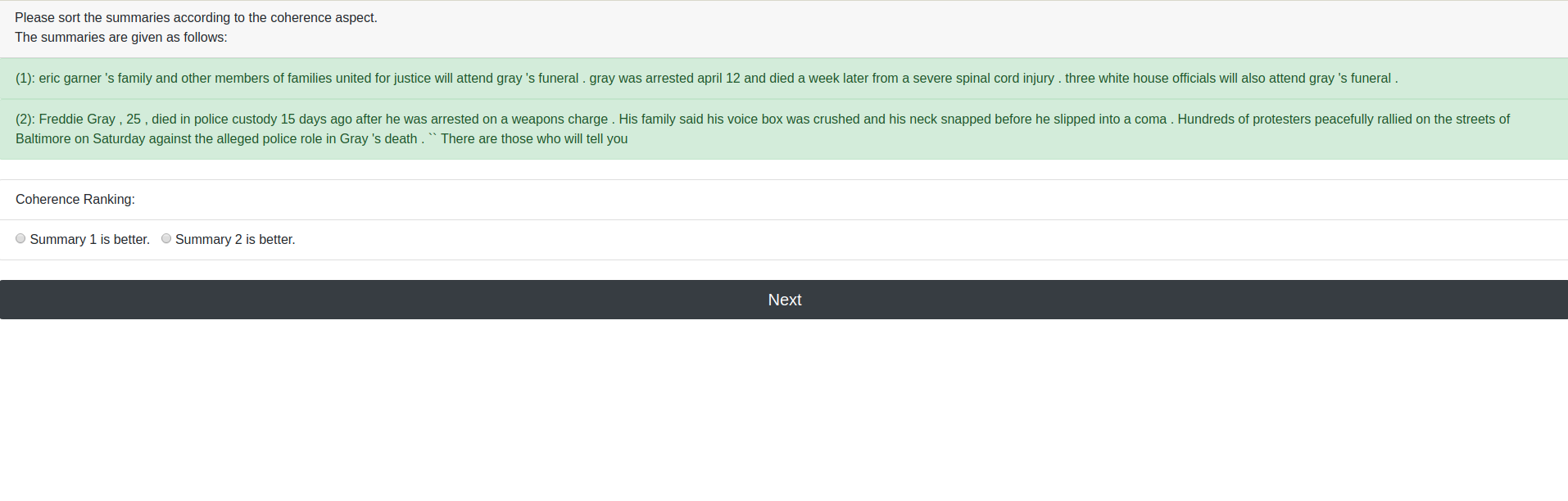}
\captionof{figure}{Human Study Interface for Abstractive Summarization}
\label{fig:humanintersum}
\end{center}
\end{minipage}

\end{document}